\documentclass[letterpaper, 10pt, conference]{ieeeconf}  

\IEEEoverridecommandlockouts                              
\usepackage[utf8]{inputenc}
\usepackage[T1]{fontenc}
\usepackage{amsfonts}
\usepackage[table]{xcolor}
\definecolor{lightblue}{rgb}{0.8,0.9,0.98}
\usepackage{amssymb}
\usepackage{amsmath}
\usepackage{graphicx}
\usepackage{trimclip} 
\newsavebox{\ablationimagebox}
\newlength{\ablationbottomtrim}
\usepackage{booktabs}
\usepackage{cite}
\usepackage{url}
\usepackage{tabularx}
\usepackage{caption}  

\makeatletter
\let\NAT@parse\undefined
\makeatother
\usepackage[colorlinks=true,linkcolor=blue,citecolor=blue,urlcolor=black]{hyperref}
\title{\LARGE \bf
Preserving Operator Intent Under Actuation Constraints: Teleoperation of Manipulators with Coupled Transmissions
}

\author{
Xuhan Miao$^{1,*}$,
Zijin Han$^{1,*}$,
Yutong Shen$^{1}$,
Haidong Huang$^{2,6, **}$,
Shanliushui Gao$^{1}$,\\
Lei Zhang$^{3,4,5}$,
Xiaocong Li$^{6}$,
and Chenguang Yang$^{2,\dagger}$
\thanks{*These authors contributed equally to this work.}
\thanks{**This work was initiated and led by Haidong Huang.}
\thanks{$\dagger$Corresponding author: Chenguang Yang.}
\thanks{$^{1}$Beijing University of Technology, Beijing, China.}
\thanks{$^{2}$The Hong Kong Polytechnic University, Hong Kong, China.}
\thanks{$^{3}$SynnoverAI.}
\thanks{$^{4}$Technical University of Munich, Munich, Germany.}
\thanks{$^{5}$University of Hamburg, Hamburg, Germany.}
\thanks{$^{6}$Eastern Institute of Technology, Ningbo, China.}
}

\begin{document}

\makeatletter
\let\@oldmaketitle\@maketitle
\renewcommand{\@maketitle}{%
  \@oldmaketitle%
  \par\vspace{1em}%
  \setcounter{figure}{0}%
  \begin{center}%
    \includegraphics[width=\linewidth, height=0.3\textheight, keepaspectratio]{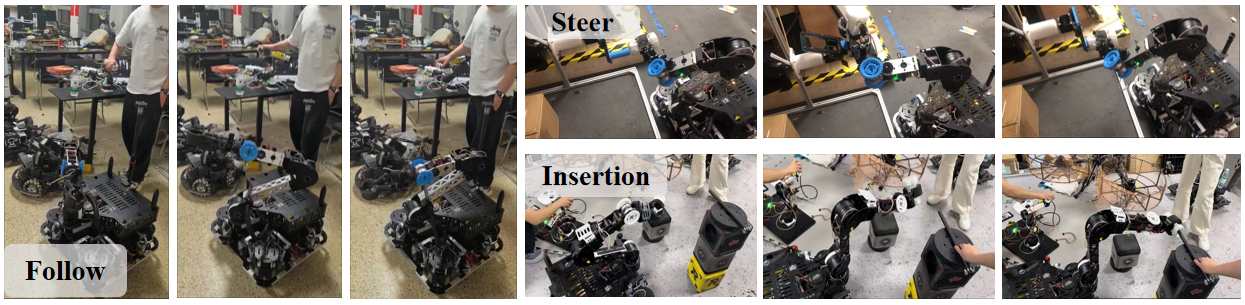}%
    \\[0.5em]%
    \captionof{figure}{%
    We propose an actuator-constraint-aware teleoperation method for manipulators with coupled transmissions that preserves the operator’s multi-DoF motion intent under gravitational loads and motor saturation, enabling continuous and predictable behavior during constraint activation and recovery.
    }%
    \label{fig:teaser}%
  \end{center}%
  \vspace{-1em}%
}
\makeatother

\maketitle
\thispagestyle{empty}
\pagestyle{empty}

\begin{abstract}

Teleoperation enables intuitive control of remote engineering robots for precision tasks. Existing direct joint-mapping methods have two limitations: independent motor saturation disrupts pitch–roll coordination in differential wrists, while fixed mappings struggle with abrupt tracking changes caused by gravitational loads and actuation constraints. We propose an actuator-constraint-aware leader–follower teleoperation method that jointly models leader gravity compensation, leader–follower joint mapping, and follower differential-wrist control. Specifically, we perform coordinated saturation in motor space according to the available gravity margin and integrate continuous angle unwrapping, feedback-state monitoring, and feasible-reference updates. This preserves the operator’s motion direction, reduces coordination distortion, and suppresses recovery transients during saturation or brief communication failures. Real-robot experiments demonstrate smoother wrist trajectories and more continuous recovery than direct mapping.

\end{abstract}

\section{INTRODUCTION}

Teleoperation enables humans to operate robots in hazardous, confined, or contact-uncertain environments, and has become an important interface for remote manipulation and demonstration collection~\cite{c1,c2}. Compared with keyboards, joysticks, and vision-only interfaces, a leader manipulator directly expresses the operator's joint motion, reducing cognitive load caused by action planning and coordinate transformation. Recent low-cost leader--follower systems support dexterous manipulation and imitation learning, but commonly rely on independent joint mapping or end-effector tracking, without explicitly modeling transmission coupling, configuration-dependent gravity, and actuator saturation~\cite{c2,c3}.

This issue is particularly pronounced in the engineering manipulator studied here. The leader and follower exchange joint angles, wrist commands, and gripper states through a custom serial protocol. The follower combines heterogeneous motors with a differential wrist, and uses proportional mapping, PID current control, gravity compensation, encoder unwrapping, and feedback timeout protection. In the existing implementation, the differential wrist drives pitch and roll through sum--difference motor commands, followed by independent motor saturation~\cite{c4}. When one motor saturates, the realized pitch--roll ratio and motion direction deviate from the operator command. Moreover, the distributed treatment of gravity compensation, angle continuity, and feedback loss does not provide a unified model of the available actuation margin. Consequently, fixed mapping cannot capture configuration-dependent capability, while temporarily infeasible commands may accumulate and cause abrupt recovery motions.

We propose an actuator-constrained teleoperation method for manipulators with coupled transmissions. The method maps transmission coupling, gravity demand, actuator limits, and feedback state into a unified actuator-space model. Operator commands are then projected onto the feasible set while preserving local motion direction and pitch--roll coordination as much as possible. A continuous feasible-reference update further prevents command accumulation during saturation or short communication dropouts. The resulting design extends the existing leader--follower hardware and real-time control pipeline, while addressing motion distortion caused by coupled actuation and gravity-dependent capability.

The contributions are threefold:
\begin{enumerate}
    \item We formulate the motion distortion induced by differential transmission and actuator-level saturation, and characterize how gravity compensation changes the available actuation margin.
    \item We propose an actuator-space intent-preserving command projection that integrates gravity compensation, joint mapping, and PID current control, thereby reducing coordination distortion near saturation.
    \item We introduce continuous feasible-reference management using encoder unwrapping and online feedback monitoring, which suppresses error accumulation and recovery transients after temporary infeasibility.
\end{enumerate}

\section{Related Work}

\subsection{Immersive and Digital-Twin Teleoperation}

Immersive teleoperation aims to reduce the cognitive gap between the operator and the remote robot by providing a spatially consistent visual or digital representation of the robot and its environment. Recent systems use virtual reality, stereoscopic feedback, or digital twins to improve viewpoint control, robot compatibility, and operator intuitiveness~\cite{c5,c6,c7}. These methods mainly improve the human--robot interface, but the generated commands are often passed to the robot through direct joint tracking or Cartesian control. Consequently, they do not explicitly address how coupled transmissions, gravity-dependent load, and motor saturation distort coordinated motions at the actuator level. Our method complements immersive interfaces with an actuator-space feasibility layer that preserves coordinated operator motion when the follower reaches its execution limits.

\subsection{Learning-Based Teleoperation and Shared Autonomy}

Learning-based teleoperation uses imitation learning, reinforcement learning, or learned intent recognition to reduce operator workload and support autonomous assistance. Recent work learns mode switching from human intent~\cite{c8}, combines teleoperation with open-world shared control~\cite{c6}, and uses large-scale teleoperated data to improve manipulation policies~\cite{c9}. These methods can handle complex task-level behavior, but their performance depends on training-data coverage and learned policies do not inherently guarantee safe execution under unseen actuator constraints. Moreover, learning-based assistance may modify the operator's command without explicitly preserving its instantaneous multi-DoF coordination. Our model-based method operates before policy execution and preserves the operator's local motion direction under gravity and actuator saturation without requiring task-specific training data.

\subsection{Optimization-Based and Constraint-Aware Robot Control}

Constraint-aware control regulates robot commands according to collision, singularity, torque, velocity, or contact constraints. Existing approaches use control barrier functions~\cite{c10}, Cartesian impedance and null-space projection, or quadratic programming to maintain safety and feasibility~\cite{c11}. These methods provide formal or numerical constraint handling, but most impose constraints in joint or Cartesian space and treat actuator saturation as a final clipping operation. Such clipping can change the coordination ratio of coupled degrees of freedom, while accumulated infeasible references may produce abrupt recovery motions. Our method formulates the coupled differential wrist and gravity-compensated actuator margins directly in actuator space, then performs coordinated command projection and continuous feasible-reference updates to reduce motion distortion and recovery transients.

\section{Method}

\subsection{Problem Formulation and System Model}

Let $q_m\in\mathbb{R}^{n_m}$ and $q\in\mathbb{R}^{n}$ denote the joint configurations of the leader and follower manipulators, respectively. The leader displacement is calibrated by
\begin{equation}
\Delta q_{\mathrm{map}}
=
S\left(q_m-q_{m,0}\right),
\qquad
\xi_h=
\begin{bmatrix}
K_x & 0\\
0 & K_R
\end{bmatrix}
\frac{\Delta q_{\mathrm{map}}}{\Delta t},
\label{eq:leader_mapping}
\end{equation}
where $S$ contains the joint direction and scale factors, while $K_x$ and $K_R$ convert the mapped motion into translational and rotational velocities.

The follower kinematics are described by the standard Denavit--Hartenberg model:
\begin{equation}
{}^{0}T_n(q)
=
\prod_{i=1}^{n}
\left[
R_z(q_i)T_z(d_i)T_x(a_i)R_x(\alpha_i)
\right]
=
\begin{bmatrix}
R(q) & p(q)\\
0 & 1
\end{bmatrix},
\label{eq:dh_model}
\end{equation}
where $p(q)$ and $R(q)$ denote the end-effector position and orientation. The corresponding spatial velocity is
\begin{equation}
\xi=
\begin{bmatrix}
v\\
\omega
\end{bmatrix}
=
J(q)\dot q,
\label{eq:jacobian}
\end{equation}
where $J(q)$ is the geometric Jacobian.

\begin{figure*}[t]
    \centering
    \includegraphics[width=0.8\textwidth]{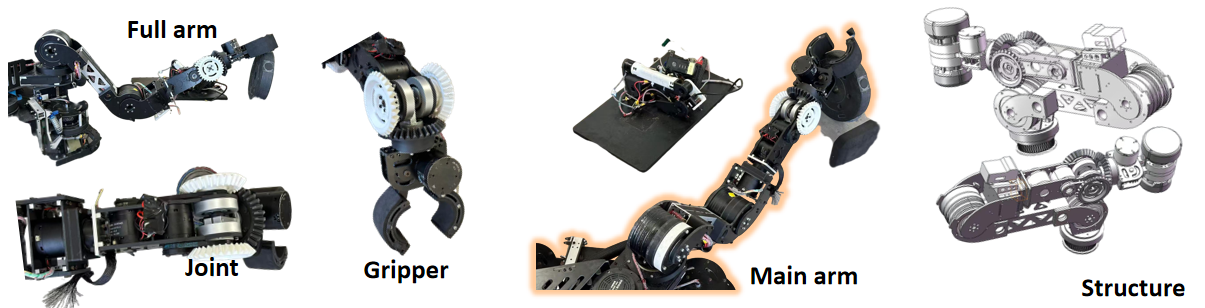}
    \caption{The robotic arm is a serial multi-DOF biomimetic manipulator that adopts an anthropomorphic joint layout with multi-stage transmission integration. Its kinematic chain consists of an end-effector gripping mechanism, a wrist rotary joint, an elbow pitch joint, and a shoulder pitch joint connected in series. All actuation units are DM (Damiao) series servo motors, and independent or coupled motion of each degree of freedom is realized through gear transmission, synchronous timing-belt transmission, and differential coupling mechanisms. Specifically, the end-effector gripping mechanism is driven by a single DM4310 motor through a gear-meshing opposed transmission, enabling symmetric synchronous opening and closing of the two jaws with a compact form factor and low distal inertia. The wrist rotary joint employs a dual-DM4310 differential coupling drive, in which power is transmitted via two synchronous timing-belt stages to the lateral drive gears, and differential synthesis produces large-angle wrist rotation while reducing the radial dimension and optimizing the overall mass distribution. The elbow and shoulder joints serve as the primary motion and load-bearing joints, each equipped with a high-torque DM8009 servo motor. These joints adopt an electromechanically integrated mounting design, and the arm body uses a hollow truss structure, thereby reducing self-weight and improving dynamic performance while maintaining bending and torsional stiffness.}
    \label{fig:double}
\end{figure*}

For the differential wrist, let
$q_w=[q_p,q_r]^{\mathsf T}$ denote the physical pitch--roll coordinates and
$\theta_w=[\theta_L,\theta_R]^{\mathsf T}$ the motor coordinates. Their transmission relationship is
\begin{equation}
\begin{aligned}
\theta_w&=Bq_w+\theta_{w,0},\\
\dot{\theta}_w&=B\dot q_w,\\
\tau_w&=B^{\mathsf T}\tau_m,
\end{aligned}
\qquad
B=
\begin{bmatrix}
-\kappa_p & \kappa_r\\
\kappa_p & \kappa_r
\end{bmatrix},
\label{eq:differential_transmission}
\end{equation}
where $B$ is the transmission matrix. Therefore, the two physical wrist degrees of freedom share the same actuator capacity, and motor-level limits result in coupled constraints in the pitch--roll space.

\subsection{Gravity Compensation and Actuator Feasibility}

The follower dynamics are modeled as
\begin{equation}
M(q)\ddot q
+
C(q,\dot q)\dot q
+
g(q)
=
\tau_q+\tau_{\mathrm{ext}},
\label{eq:dynamics}
\end{equation}
where $M(q)$ is the inertia matrix, $C(q,\dot q)\dot q$ contains the Coriolis and centrifugal terms, $g(q)$ is the gravity torque, and $\tau_{\mathrm{ext}}$ is the external torque. The gravity vector is obtained from the potential energy:
\begin{equation}
\begin{aligned}
U(q)&=\sum_{k=1}^{N}m_k g_0 h_k(q),\\
g(q)&=\frac{\partial U(q)}{\partial q}.
\end{aligned}
\label{eq:gravity_model}
\end{equation}

For the differential wrist, the physical gravity torque is transformed into the motor space according to
\begin{equation}
\tau_{g,m}(q)
=
B^{-\mathsf T}g_w(q).
\label{eq:wrist_gravity}
\end{equation}
Let $\tau_{m,\min}$ and $\tau_{m,\max}$ denote the motor torque limits. The motion torque that remains available after gravity compensation is
\begin{equation}
\begin{aligned}
\mathcal{A}(q)
=
\Big\{
\tau_{m,\mathrm{motion}}\ \Big|\
&\tau_{m,\min}-\tau_{g,m}(q)
\leq\tau_{m,\mathrm{motion}}\\
&\leq\tau_{m,\max}-\tau_{g,m}(q)
\Big\}.
\end{aligned}
\label{eq:actuation_set}
\end{equation}
Thus, the available actuator capability varies with the robot configuration and payload.

The motor velocity constraints are
\begin{equation}
\dot\theta_{\min}
\leq B\dot q_w
\leq\dot\theta_{\max}.
\label{eq:motor_velocity_constraint}
\end{equation}
For symmetric limits and equal transmission ratios, the differential wrist satisfies
\begin{equation}
\begin{aligned}
|\dot\theta_L|&\leq\bar v,
&
|\dot\theta_R|&\leq\bar v,\\
|\kappa_p\dot q_p-\kappa_r\dot q_r|
&\leq\bar v,
&
|\kappa_p\dot q_p+\kappa_r\dot q_r|
&\leq\bar v.
\end{aligned}
\label{eq:differential_constraints}
\end{equation}
When $\kappa_p=\kappa_r$, the feasible set becomes
$|\dot q_p|+|\dot q_r|\leq\bar v/\kappa_p$, which is a coupled diamond-shaped region rather than two independent joint intervals.

\subsection{Actuator-Space Intent-Preserving Projection}

A direct inverse-kinematics command,
\begin{equation}
\dot q=J^{\dagger}(q)\xi_h,
\label{eq:direct_ik}
\end{equation}
does not guarantee joint, transmission, or actuator feasibility. We therefore introduce an execution-progress variable $\alpha\in[0,1]$, where $\alpha=1$ denotes full execution of the operator command. The feasible follower velocity is obtained from
\begin{equation}
\begin{aligned}
\min_{\dot q,\alpha}\quad
&
\left\|J(q)\dot q-\alpha\xi_h\right\|_{W}^{2}
+
\lambda_v\left\|\dot q-\dot q_{k-1}\right\|^{2}\\
&
+
\lambda_\alpha(1-\alpha)^2
\end{aligned}
\label{eq:projection_objective}
\end{equation}
subject to
\begin{equation}
\begin{aligned}
q_{\min}
&\leq q+\Delta t\dot q
\leq q_{\max},\\
\dot\theta_{\min}
&\leq B(q)\dot q
\leq\dot\theta_{\max},\\
\tau_{m,\min}
&\leq
B^{-\mathsf T}\tau_w(\dot q)
+\tau_{g,m}(q),\\
B^{-\mathsf T}\tau_w(\dot q)
+\tau_{g,m}(q)
&\leq\tau_{m,\max},\\
0&\leq\alpha\leq1.
\end{aligned}
\label{eq:projection_constraints}
\end{equation}

\begin{figure*}[t]
    \centering
    \includegraphics[width=0.8\textwidth]{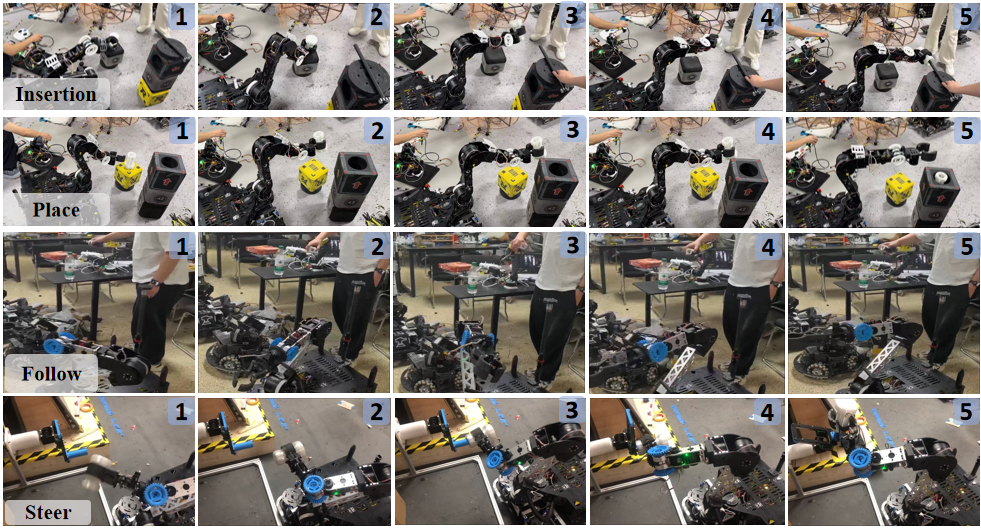}
    \caption{To verify the effectiveness and robustness of our method, we evaluate it on four tasks: Insertion, Place, Follow, and Steer. The figure illustrates the successful execution procedure for each task.}
    \label{fig:double}
\end{figure*}
The first objective term preserves the desired end-effector motion, the second suppresses command discontinuities, and the third maximizes the executable motion. To explicitly penalize directional distortion, we define
\begin{equation}
P_{\parallel}
=
\frac{\xi_h\xi_h^{\mathsf T}}
{\xi_h^{\mathsf T}\xi_h+\varepsilon},
\qquad
P_{\perp}=I-P_{\parallel}.
\label{eq:direction_projectors}
\end{equation}
The direction-prioritized objective is
\begin{equation}
\begin{aligned}
\min_{\dot q,\alpha}\quad
&
\left\|P_{\perp}J(q)\dot q\right\|^{2}\\
&
+
\lambda_{\parallel}
\left\|
P_{\parallel}J(q)\dot q-\alpha\xi_h
\right\|^{2}\\
&
+
\lambda_v
\left\|\dot q-\dot q_{k-1}\right\|^{2}.
\end{aligned}
\label{eq:direction_objective}
\end{equation}
The first term suppresses motion components orthogonal to the operator command, while the second term preserves the command component parallel to the desired direction. Consequently, when the follower cannot execute the full command, the controller first reduces the execution speed and only then modifies the motion direction.

For the differential wrist, the local projection can be written as
\begin{equation}
\begin{aligned}
\min_{\dot q_w,\alpha}\quad
&
\left\|
\dot q_w-\alpha\dot q_{w,h}
\right\|_{W_w}^{2}
+
\lambda_w
\left\|
\dot q_w-\dot q_{w,k-1}
\right\|^{2}\\
\mathrm{s.t.}\quad
&
\dot\theta_{\min}
\leq B\dot q_w
\leq\dot\theta_{\max}.
\end{aligned}
\label{eq:wrist_projection}
\end{equation}
This prevents independent motor clipping from changing the relative pitch--roll motion requested by the operator.

\subsection{Continuous Feasible Reference}

If the raw leader command is directly integrated into the follower target, infeasible motion accumulates as position error and may generate a catch-up motion after the actuator becomes available. We therefore update the follower reference using only the optimized feasible velocity:
\begin{equation}
q_{\mathrm{ref},k+1}
=
q_{\mathrm{ref},k}
+
\Delta t\,\dot q_k^{*},
\label{eq:reference_update}
\end{equation}
where $\dot q_k^{*}$ is the solution of the constrained projection problem.

For periodic encoder variables, the continuous state is updated by
\begin{equation}
\tilde q_k
=
\tilde q_{k-1}
+
\operatorname{wrap}
\left(q_k-q_{k-1}\right),
\label{eq:angle_unwrap}
\end{equation}
where $\operatorname{wrap}(\cdot)$ returns the shortest angular difference. This state is used for velocity estimation, continuous wrist-roll accumulation, and reference generation.

Let $\mathcal{V}_k\in\{0,1\}$ indicate whether the feedback is valid. If the feedback is unavailable or exceeds the timeout threshold, the reference is frozen and synchronized with the measured configuration:
\begin{equation}
\mathcal{V}_k=0
\quad\Longrightarrow\quad
\dot q_k^{*}=0,
\qquad
q_{\mathrm{ref},k}=q_{\mathrm{real},k}.
\label{eq:feedback_safety}
\end{equation}
After valid feedback is restored, the reference is restarted from the current measured configuration rather than from an unexecuted historical command. This prevents communication loss or temporary saturation from producing a delayed recovery transient.

\subsection{Low-Level Joint and Motor Control}

The feasible reference is tracked using cascaded position--velocity control with gravity feedforward:
\begin{equation}
\tau_q
=
K_p
\left(q_{\mathrm{ref}}-q\right)
+
K_d
\left(\dot q_{\mathrm{ref}}-\dot q\right)
+
g(q).
\label{eq:joint_control}
\end{equation}
For the differential wrist, the physical pitch--roll torque is
\[
\tau_w=
\begin{bmatrix}
\tau_p\\
\tau_r
\end{bmatrix},
\]
and the corresponding motor torque and current are
\begin{equation}
\begin{aligned}
\tau_m
&=
B^{-\mathsf T}\tau_w
+
\tau_{g,m}(q),\\
i_m
&=
K_t^{-1}\tau_m,
\end{aligned}
\label{eq:motor_control}
\end{equation}
where $K_t$ is the motor torque constant.

The complete control loop consists of leader measurement, calibrated mapping, kinematic transformation, gravity and transmission modeling, actuator-space feasibility projection, continuous reference update, and joint--motor control. When sufficient actuation is available, $\alpha$ remains close to one and the follower tracks the leader directly. When gravity load or actuator saturation reduces the available capability, $\alpha$ decreases while the operator's local motion direction and differential-wrist coordination are preserved.

\begin{figure*}[t]
    \centering
    \includegraphics[width=0.8\textwidth]{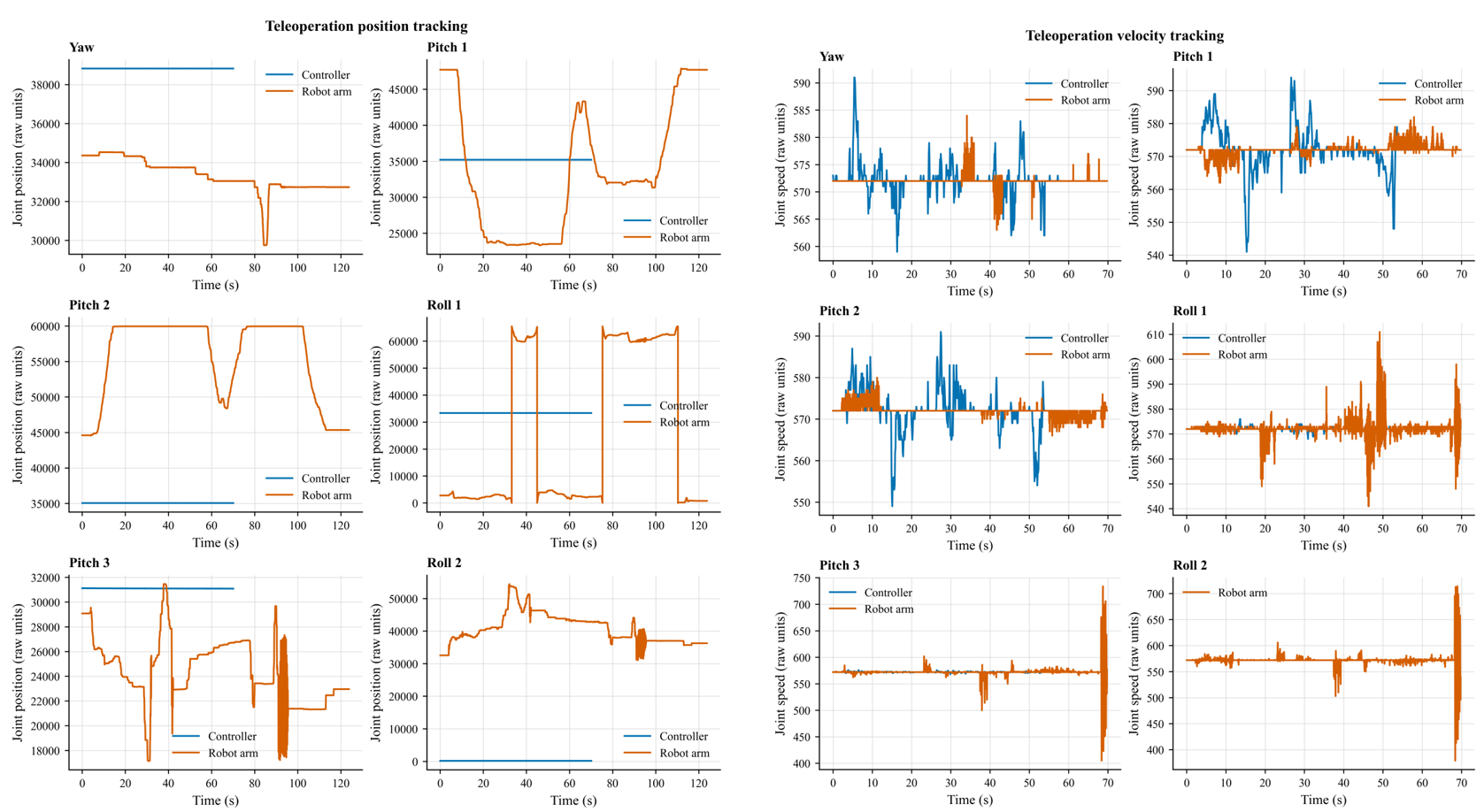}
    \caption{To validate the overall performance of our method, we recorded and compared in real time the indicators of the controller and the on-vehicle arm.}
    \label{fig:re1}
\end{figure*}

\section{Experiment}
\subsection{Setup}
We validated the proposed master, slave teleoperation control method on a real master, slave teleoperation system. Joint position and velocity signals were collected at both the controller and the slave manipulator. Each record contains a timestamp and multiple joint channels, enabling comparison between operator input and the actual execution state of the manipulator.

The experimental signals contain six joint channels: Yaw, Pitch 1, Pitch 2, Roll 1, Pitch 3, and Roll 2. Since some channel names in the raw data are duplicated or inconsistent, we match each channel by its column order in the records rather than relying directly on the original names. On the controller side, the position data contain 688 samples over approximately 70.07 s, with a mean sampling rate of about 9.82 Hz; the slave manipulator position data contain 12,376 samples over approximately 123.75 s, with a mean sampling rate of about 100.01 Hz. For the velocity signals, the controller side contains 570 samples over approximately 58.00 s, with a mean sampling rate of about 9.82 Hz; the slave manipulator contains 6,977 samples over approximately 69.76 s, with a mean sampling rate of about 100.01 Hz.

This experiment aims to verify whether the proposed actuator-constraint-aware teleoperation method can maintain continuous and coordinated joint motion between controller input and actual manipulator execution. By comparing position and velocity responses at the controller and manipulator sides, we focus on tracking continuity and motion consistency under differential wrist Pitch–Roll coupling, actuator capability limits, and transient variations, thereby validating the method's effectiveness in reducing motion distortion, preserving joint coordination, and improving recovery under constrained conditions.

\subsection{Result}
As shown in Figure~\ref{fig:re1}. Experimental results show that the controller-side input signals are generally smooth, with the Yaw, Pitch 1, and Pitch 2 position channels varying by only 2--3 raw units, whereas the Roll 1 and Pitch 3 channels vary by 21 and 44 raw units, respectively, indicating pronounced wrist posture adjustments during the experiment. In the velocity data, the mean value of each controller-side channel is approximately 572 raw units, and the corresponding manipulator-side means are also close to 572 raw units, indicating that the manipulator maintains an overall motion level comparable to that of the controller. The manipulator-side velocity standard deviations for Yaw, Pitch 1, and Pitch 2 are 0.90, 1.66, and 1.26 raw units, respectively, showing small fluctuations, whereas the ranges of Pitch 3 and Roll 2 reach 329 and 335 raw units, reflecting pronounced transient responses induced by differential wrist coupling and actuator constraints. Overall, the manipulator maintains continuous motion for most of the duration without evident sustained error accumulation, demonstrating the effectiveness of the actuator-space constraint projection and continuous feasible reference update strategies.

\begin{figure}[t]
    \centering
    \begingroup
    \sbox{\ablationimagebox}{%
        \includegraphics[width=\linewidth]{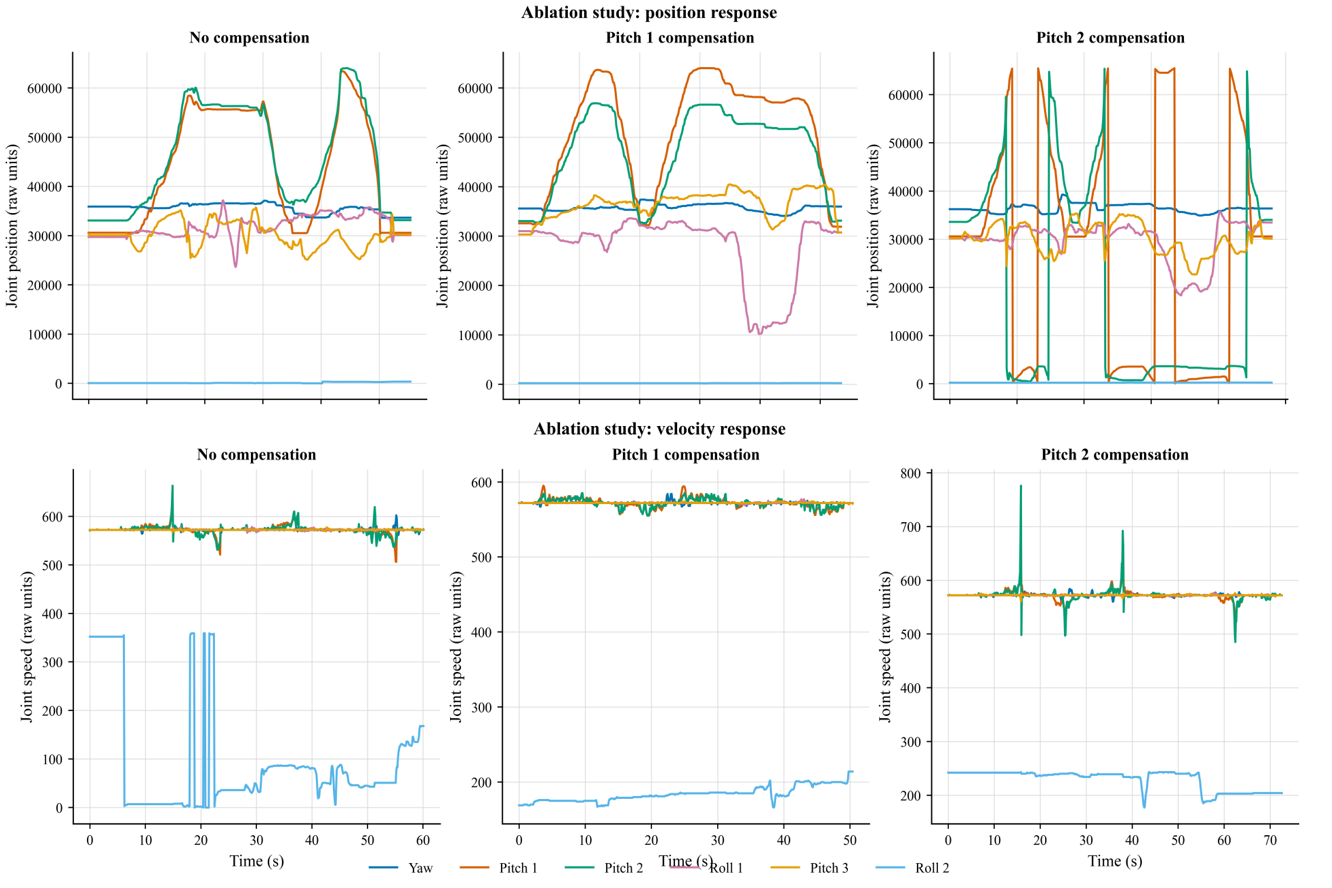}%
    }%
    \setlength{\ablationbottomtrim}{0.0461712\ht\ablationimagebox}%
    \clipbox{0pt \ablationbottomtrim{} 0pt 0pt}{%
        \usebox{\ablationimagebox}%
    }\par
    \vspace{1pt}
    {\fontsize{6}{7}\selectfont
    \noindent\makebox[\linewidth][l]{%
        \hspace*{0.188\linewidth}%
        \makebox[0pt]{Time (s)}%
        \hspace*{0.323\linewidth}%
        \makebox[0pt]{Time (s)}%
        \hspace*{0.326\linewidth}%
        \makebox[0pt]{Time (s)}%
    }\par}
    \vspace{3pt}
    \definecolor{ablationYaw}{HTML}{0072BD}
    \definecolor{ablationPitchOne}{HTML}{D95319}
    \definecolor{ablationPitchTwo}{HTML}{009E73}
    \definecolor{ablationRollOne}{HTML}{CC79A7}
    \definecolor{ablationPitchThree}{HTML}{E69F00}
    \definecolor{ablationRollTwo}{HTML}{56B4E9}
    \newcommand{\ablationlegendentry}[2]{%
        \raisebox{0.25ex}{\textcolor{#1}{\rule{1.6em}{0.65pt}}}%
        \hspace{0.4em}#2%
    }
    {\fontsize{7}{8}\selectfont
    \setlength{\tabcolsep}{0pt}%
    \begin{tabular*}{\linewidth}{@{\extracolsep{\fill}}lll@{}}
        \ablationlegendentry{ablationYaw}{Yaw} &
        \ablationlegendentry{ablationPitchOne}{Pitch 1} &
        \ablationlegendentry{ablationPitchTwo}{Pitch 2} \\[2pt]
        \ablationlegendentry{ablationRollOne}{Roll 1} &
        \ablationlegendentry{ablationPitchThree}{Pitch 3} &
        \ablationlegendentry{ablationRollTwo}{Roll 2}
    \end{tabular*}\par}
    \endgroup
    \caption{Ablation study of gravity compensation. Joint position (top) and velocity (bottom) responses are shown for no compensation, Pitch 1 compensation, and Pitch 2 compensation. The shared legend identifies the six joint channels.}
    \label{fig:re5}
\end{figure}

\subsection{Ablation Study}
\subsubsection{Settings}
To verify the impact of gravity compensation on actuator available capacity and coupled joint motion, we designed three ablation experiments: no gravity compensation, compensation for the Pitch 1 joint only, and compensation for the Pitch 2 joint only. All three experiments kept the same master-slave joint mapping, differential wrist control, actuator constraints, and continuous feasible reference update strategy, with only the gravity compensation configuration varied. Since gravity compensation directly affects the residual torque margin in motor space, different compensation strategies lead to different joint tracking capabilities and transient responses. By comparing the joint position and velocity profiles across the three experiments, we focus on analyzing the effects of gravity compensation on load-dependent joint stability, motion continuity, differential wrist coordination, and recovery from constrained states, thereby validating the necessity of the gravity compensation module within the proposed actuator-constraint-aware teleoperation framework.
\subsubsection{Result}
As shown in Figure~\ref{fig:re5}. Based on the ablation experimental data, the full method exhibits a clear advantage in velocity response stability. The average standard deviation of the manipulator velocity signals decreased from 22.41 without gravity compensation to 5.91 with the full method, corresponding to an overall fluctuation reduction of approximately 73.6\%. In particular, under the no-compensation condition, the standard deviation of the Roll 2 channel reached 110.68, whereas the full method reduced it to 14.39, a decrease of approximately 87.0\%, indicating that gravity compensation and actuator constraint handling can effectively suppress severe transient responses in the differential wrist. When compensating only Pitch 1, the average velocity standard deviation across the six channels was further reduced to 3.99, approximately 82.2\% lower than that without compensation, with fluctuations in Pitch 2 and Roll 2 decreasing by approximately 47.4\% and 91.2\%, respectively. In contrast, when compensating only Pitch 2, the average standard deviation was 7.58, and the fluctuation in the Pitch 2 channel actually increased from 9.84 to 14.22. This suggests that compensating a single load-bearing joint alone may improve local actuation capability but cannot fully resolve the overall coordination problem in the coupled wrist. Overall, the ablation results validate the effectiveness of gravity compensation in improving joint motion stability, suppressing velocity spikes, and enhancing differential wrist coordination, and further demonstrate the necessity of jointly designing gravity compensation, actuator-space constraints, and continuous feasible reference updates.

\section{Conclusion}
We proposed an actuator-constraint-aware teleoperation method that unifies leader gravity compensation, joint mapping, and differential-wrist control in actuator space, preserving operator intent under gravitational loads and motor saturation. Coordinated command projection, continuous angle unwrapping, and feasible-reference updates reduce coordination distortion and recovery transients. Real-robot experiments and ablation studies confirmed smoother wrist trajectories and improved velocity stability, with the full method lowering the average velocity standard deviation by 73.6\% over the no-compensation baseline. These results validate the necessity of jointly designing gravity compensation, actuator-space constraints, and continuous feasible reference updates.

\end{document}